\documentclass{article}
\usepackage{CJKutf8}
\usepackage{microtype}
\usepackage{graphicx}
\usepackage{subcaption}
\usepackage{booktabs} 
\usepackage{tcolorbox}
\usepackage{hyperref}

\usepackage[preprint]{icml2026}

\usepackage{amsmath}
\usepackage{amssymb}
\usepackage{mathtools}
\usepackage{amsthm}
\usepackage[normalem]{ulem}
\useunder{\uline}{\ul}{}
\usepackage[table,xcdraw]{xcolor}

\usepackage[capitalize,noabbrev]{cleveref}

\theoremstyle{plain}

\theoremstyle{definition}

\theoremstyle{remark}

\usepackage[textsize=tiny]{todonotes}

\icmltitlerunning{L-MAD: A Systematic Evaluation of Multi-Agent Debate Structures in Legal Reasoning}

\begin{document}

\twocolumn[
  \icmltitle{L-MAD: A Systematic Evaluation of Multi-Agent Debate Structures \\ in Legal Reasoning}



  \icmlsetsymbol{equal}{*}

    \begin{icmlauthorlist}
        \icmlauthor{Tan-Minh Nguyen}{equal,jaist}
        \icmlauthor{Hoang-Trung Nguyen}{equal,vnu}
        \icmlauthor{Huu-Dong Nguyen}{equal,vnu}
        \icmlauthor{Truong Dinh Do}{jaist}
        \icmlauthor{Thi-Hai-Yen Vuong}{vnu}
        \icmlauthor{Le-Minh Nguyen}{jaist}
    \end{icmlauthorlist}
    
    \icmlaffiliation{jaist}{Japan Advanced Institute of Science and Technology (JAIST), Ishikawa, Japan}
    \icmlaffiliation{vnu}{VNU University of Engineering and Technology, Hanoi, Vietnam}
    
    \icmlcorrespondingauthor{Le-Minh Nguyen}{nguyenml@jaist.ac.jp}

  \icmlkeywords{Machine Learning, ICML}

  \vskip 0.3in
]



\printAffiliationsAndNotice{}  

\begin{abstract}
While multi-agent debate (MAD) frameworks have shown significant potential in general reasoning, their effectiveness in highly structured, knowledge-heavy legal domains remains under-explored. In this work, we introduce the Legal Multi-Agent Debate (L-MAD) framework to systematically evaluate different debate structures and aggregation methods within Legal Textual Entailment. By assigning distinct expert personas to multiple agents, L-MAD improves upon strong single-agent baselines by up to 8\%. Furthermore, analyzing how debate scales reveals a clear trade-off: increasing the agent population reduces inconsistency and improves accuracy, whereas extending discussion rounds induces a detrimental \textit{over-deliberation drift} where agents reinforce each other's mistakes. Ultimately, our findings outline the practical boundaries and safety margins of deploying collaborative multi-agent systems in high-stakes legal reasoning environments.
\end{abstract}

\section{Introduction}

Large language models (LLMs) show remarkable capabilities in general reasoning, but they frequently struggle in high-stakes, knowledge-heavy domains that demand strict logic. Legal Textual Entailment (LTE) \cite{aoki2022data,bilgin2024exploring} is a challenging test for these models, requiring them to determine whether specific legal statutes apply to a given factual scenario \cite{aletras2016predicting,zhong-etal-2020-nlp}. While specialized models like LegalBERT \cite{chalkidis-etal-2020-legal} and Lawformer \cite{xiao2021lawformer} outperform generalist baselines, standard single-model inference is still limited by a model's basic reasoning capacity and its tendency to make factual or logical errors.

\begin{figure}[t]
  \begin{center}
    \centerline{\includegraphics[width=\columnwidth]{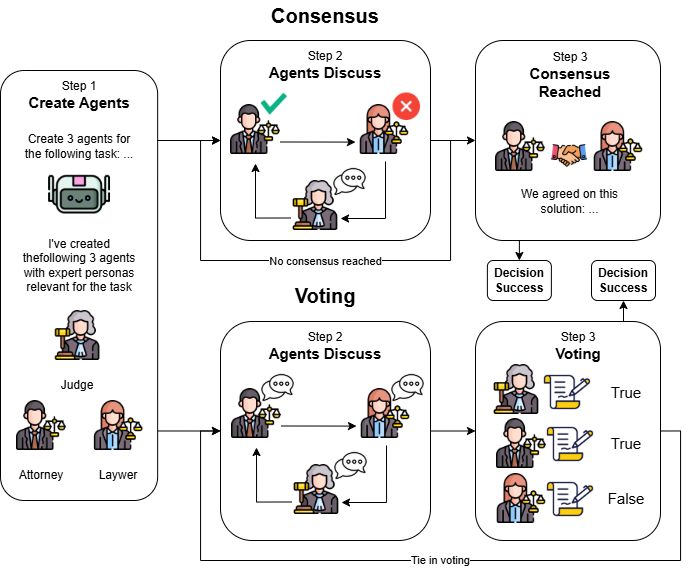}}
    \caption{
      Illustration of decision-making protocols used in this study.
    }
    \label{fig:framework}
  \end{center}
\end{figure}

Multi-Agent Debate (MAD) frameworks offer a promising way to overcome these limitations by allowing models to spend more computing power during inference \cite{nie-etal-2020-adversarial,choi2026debate}. By assigning distinct expert personas to multiple LLM agents, these frameworks let models iterate and refine their arguments to arrive at a better final answer \cite{kaesberg2025voting}. However, how these debate frameworks perform in highly structured, rule-bound fields like law remains underexplored. While open-ended debate typically helps in general tasks, it introduces unique challenges in legal reasoning. The trade-offs between forcing agents to agree versus letting them vote independently—and the risk of errors multiplying over several rounds—have not yet been systematically analyzed.

To bridge this gap, we introduce Legal Multi-Agent Debate (L-MAD), a framework designed to analyze different multi-agent debate structures under strict legal rules (Figure~\ref{fig:framework}). Unlike prior work that mainly focuses on connecting LLMs to external tools, L-MAD provides a controlled environment to evaluate core aggregation methods. Specifically, we compare \textit{forced consensus} strategies (where agents debate until they agree) against \textit{voting} approaches (where agents maintain independent reasoning paths).

Our empirical results show that L-MAD significantly improves performance, outperforming strong single-agent baselines by up to 8\% in some settings. Crucially, we find that the best debate strategy depends heavily on the underlying model's capability. Forcing a consensus works well with highly capable models (e.g., 30B+ parameters) by encouraging constructive debate, whereas independent voting protects smaller models (e.g., 8B parameters) from superficial agreement. Furthermore, exploring how debate scales reveals a clear trade-off: adding more agents to the population reduces inconsistency and improves accuracy, but adding more discussion rounds causes \textit{over-deliberation drift}—a behavior where agents echo and validate each other's mistakes, ultimately degrading performance.

In summary, our core contributions are:
\begin{itemize}
    \item A systematic evaluation of multi-agent debate frameworks for decision-making in high-stakes, knowledge-dense legal domains.
    \item A clear analysis of the scaling trade-offs between increasing the number of participating agents versus extending the number of debate rounds.
    \item Actionable insights into the boundaries and safety margins of deploying multi-agent LLM systems in collaborative legal reasoning environments.
\end{itemize}

\section{Preliminaries}
\subsection{Legal Textual Entailment}
We employ Legal Textual Entailment to assess the ability of LLMs to accurately deduce legal outcomes from factual descriptions and statutory rules. By emphasizing subtle linguistic and doctrinal distinctions, this task rigorously evaluates whether a model possesses a genuine comprehension of legal theory. An illustrative example of the task (official translation from Japanese) is presented below:

\begin{quote}
\textbf{Premise:} \textit{(Juridical Acts by Adult Wards under Guardianship) Article 9} \\
A juridical act performed by an adult ward is voidable; provided, however, that this does not apply to the purchase of daily necessities or to any other act involved in day-to-day life.\smallskip

\textbf{Hypothesis:} If a person under adult guardianship purchases daily necessities without the consent of the adult guardian, the adult guardian may cancel the contract involving the purchase.\smallskip

\textbf{Entailment:} \textsc{No}
\end{quote}

More formally, let $\mathcal{S} = \{S_1, S_2, \dots, S_n\}$ denote a set of relevant statutory articles and let $Q$ represent a legal hypothesis. The Legal Textual Entailment task can be cast as a binary classification problem mapping the tuple $(\mathcal{S}, Q)$ to a label $y \in \{1, 0\}$, corresponding to $Entails(\mathcal{S}, Q)$ and $Entails(\mathcal{S}, \neg Q)$, respectively.

The primary evaluation metric for this task is classification accuracy. Given that the benchmark datasets are rigorously balanced between positive and negative instances, a random guessing baseline or a naive uniform-prediction heuristic yields an expected accuracy of approximately $0.50$.


\subsection{Prompting Techniques}
\paragraph{Standard Prompting and In-Context Learning.}
Standard prompting elicits an answer $a$ from a model parameterized by $\theta$ given a query $q$, formally modeling the conditional distribution $P_\theta(a \mid q)$. In-context learning \cite{brown2020language} enhances this generation process by conditioning the language model on a small set of exemplars $\mathcal{D} = \{d_1, d_2, \dots, d_n\}$, effectively modeling $P_\theta(a \mid \mathcal{D}, q)$ without requiring parameter updates.

\paragraph{Legal Reasoning CoT (IRAC).}
Eliciting intermediate reasoning steps significantly improves the proficiency of LLMs on complex cognitive tasks \cite{wei2022chain}. Standard CoT extends the in-context learning paradigm by augmenting each exemplar with a rationale $r_i$, such that $d_i = (q_i, r_i, a_i)$. The complexity and structural rigor of $r_i$ directly correlate with downstream performance \cite{fu2022complexity}. In the juridical domain, \citet{yu2022legal} specialized this approach via the IRAC framework—a canonical legal reasoning structure comprising four discrete stages: Issue (identifying the legal controversy), Rule (retrieving the governing statute), Application (mapping the rule to the factual matrix), and Conclusion (deriving the final legal verdict).

\paragraph{Self-Consistency.}
To mitigate the stochasticity of single-path decoding, self-consistency \cite{wang2022self} samples a diverse set of reasoning paths and marginalizes over them to find the most robust answer. By drawing multiple independent completions from the model, the final prediction $a^*$ is obtained via majority voting, defined as $a^* = \arg\max_{a} \sum_{i} \mathbb{I}(a_i = a)$. While this ensemble-like method effectively increases the likelihood of finding the correct reasoning pathway, it remains upper-bounded by the native inferential limits of the base model.

\subsection{Multi-Agent Debate for Decision-Making}
Multi-Agent Debate is a collaborative framework in which multiple language-model agents engage in structured interaction—typically in the form of iterative exchanges or discussions —to solve tasks such as question answering or text generation. In a typical MAD protocol, each agent independently generates an initial response and then engages in a series of debate rounds. At round $t$, an agent receives the original question along with responses from its peers at round $t - 1$, prompting the model to update its response accordingly. This iterative process is designed to leverage diverse reasoning paths and peer wisdom, potentially enhancing the overall decision quality. After all rounds of debate, the final answer is typically derived through an aggregation mechanism, such as majority voting.

\section{Methodology}

In this section, we formalize the architecture of our multi-agent framework, detailing the deliberation topology, the prompt-based response conditioning, and the decision aggregation protocols.

\subsection{Framework Configuration and Topology}
Our multi-agent debate framework is parameterized by three foundational elements: a deliberation topology, a decision protocol, and an agent-specific response generator. We initialize each debate session with three distinct, automatically generated expert personas, adapting the methodology of \citet{kim2024persona}. Empirical evidence from \citet{yin-etal-2023-exchange} demonstrates that a tripartite configuration offers an optimal theoretical trade-off between viewpoint diversity and computational overhead. By instantiating distinct personas, we effectively inject domain-specific inductive biases into the reasoning process, ensuring a robust exploration of the solution space.

During deliberation, agents exchange information over discrete communication rounds. In each round, agents operate asynchronously, attending to the broadcasted messages of their peers from previous rounds based on a sliding context window (restricted to the two most recent turns). This truncated context serves as an attention regularization mechanism to limit context dilution and prevent catastrophic drift from the original hypothesis. The exact termination condition of the debate is strictly governed by the chosen decision protocol. For consensus protocols, agents explicitly signal agreement or disagreement per turn, iteratively refining their stance until the threshold is met. For voting protocols, agents communicate for a fixed horizon (e.g., three rounds) before casting terminal votes.

\subsection{Response Generation and Decision Protocols}
The semantic structure of the agent prompt heavily influences the trajectory of multi-agent deliberation. We employ a \textit{Simple Response Generator} as our default template, which conditions agents to evaluate preceding arguments neutrally and without adversarial bias, sustaining constructive dialogue while accommodating the persona formulation.

To aggregate heterogeneous reasoning paths into a final output, we implement and evaluate two distinct classes of decision-making protocols:

\textbf{Consensus-Based Protocols.} These protocols enforce active negotiation, driving agents to mathematically converge on a unified hypothesis. A final decision is formalized only when a predefined threshold of inter-agent agreement is satisfied. We define three progressive thresholds: majority ($>50\%$ agreement), supermajority ($>66\%$ agreement), and unanimity ($100\%$ agreement). Building upon the standard majority-vote mechanism of \citet{yin-etal-2023-exchange}, we employ a hybrid consensus approach that demands unanimity for initial convergence while defaulting to a supermajority threshold to prevent deadlock at the turn limit.

\textbf{Voting-Based Protocols.} In contrast to forced consensus, voting protocols decouple the deliberation phase from the final aggregation step, permitting agents to maintain isolated, parallel reasoning tracks throughout the debate. Following the communication horizon, the protocol halts argumentation and prompts a discrete voting mechanism. In the event of a tie, the framework triggers a single supplementary deliberation round, followed by a tie-breaking vote. Inspired by \citet{yang2024llm}, we utilize a ranked-choice voting scheme wherein each agent issues a strict preference ordering over the candidate solutions (e.g., Borda count equivalent). The hypothesis maximizing the aggregated rank score is established as the final legal entailment.

\section{Experiment}

\subsection{Benchmarks}

\begin{table}[t]
\caption{Statistics of COLIEE benchmarks for three years.}
\label{tab:data_stats}
\small
\begin{tabular}{lccc}
\toprule
Data                          & \textbf{R07} & \textbf{R06} & \textbf{R05}    \\
\midrule
\# samples                    & 82 & 73 & 109    \\
avg. \# tokens per hypothesis & 93.18 & 96.04 & 102.34 \\
avg \# tokens per premise     & 231.48 & 232.51 & 283.76 \\
avg. \# relevant articles     & 1.10 & 1.26 & 1.19 \\
\bottomrule
\end{tabular}
\end{table}

For the experiments, we used the training and test datasets provided in COLIEE 2026 Task 4. Although no model fine-tuning was performed in this study, the training data were utilized to select examples for few-shot prompting.

To analyze year-by-year performance variation, we used evaluation data from three years, spanning from 2024 to 2026. The COLIEE 2026 formal run uses the 2025 test subset, while our methods comparisons are reported as a post-hoc controlled comparison over 2024-2025.
Problem IDs in the dataset follow the Japanese era naming convention, for example, R05 to 2023, and so on up to R07 for 2025.
Each sample in the training and test datasets consists of the following information: Relevant article, Query, Answer (Y: correct/N: incorrect). The statistics of benchmarks are reported in Table \ref{tab:data_stats}.

\subsection{Metric}
We used accuracy as the evaluation metric to assess the performance of each method. The model outputs were parsed from the fixed output labels: ``Answer:Y'' (Correct/True) and ``Answer:N'' (Incorrect/False). Accuracy was computed by comparing the parsed label with the ground truth label.

\subsection{Baselines and Implementation Details}
We compare L-MAD against various prompting baselines, including Zeroshot, Legal theory chain-of-thought (IRAC) \cite{yu2022legal}, Fewshot, and Self-Consistency \cite{wang2022self}. 

The LLMs used for the experiments are Qwen3-8/32B, Qwen3-30B-A3B \cite{qwen3}, Llama3.1-8B \cite{dubey2024llama}, all instruction following versions. All experiments are conducted on a single A6000 GPU with 49GB of memory. 
For implementation, we set the temperature to 0.1 for generating diverse reasoning paths. All models are loaded in full precision \texttt{bfloat16}. The self-consistency method is aggregated on three independent samples. Few-shot examples are retrieved using BM25 on the training set. The number of examples is set at 6. 

\section{Result}

\begin{table*}[t]
\centering
\caption{Task performance for two decision protocols (voting and consensus-based) on four LLM backbones. All values and standard deviations are multiplied by 100. Bold and underline indicate the highest and second-highest results per dataset. Green cells indicate that the L-MAD variant outperforms the Zeroshot baseline; red indicates it does not. Standard deviation over three runs.}
\label{tab:main_results}
\begin{tabular}{llcccc}
\hline
\textbf{Model} & \textbf{Method} & \textbf{R07} & \textbf{R06} & \textbf{R05} & \textbf{Average} \\ \toprule
Qwen3-30B-A3B & Zeroshot & $85.77_{\pm 1.41}$ & $80.37_{\pm 1.58}$ & $76.45_{\pm 1.06}$ & $80.86_{\pm 1.35}$ \\
 & IRAC & $86.59_{\pm 0.00}$ & $79.45_{\pm 0.00}$ & $75.54_{\pm 0.53}$ & $80.73_{\pm 0.18}$ \\
 & Fewshot & $89.02_{\pm 1.22}$ & $80.82_{\pm 0.95}$ & $80.73_{\pm 1.15}$ & $83.52_{\pm 1.11}$ \\
 & Self-Consistency & 86.59 & 79.45 & 76.15 & 80.73 \\
 & \cellcolor[HTML]{EFEFEF}L-MAD-consensus & \cellcolor[HTML]{D4EDDA}\textbf{95.12$_{\pm 2.58}$ } & \cellcolor[HTML]{D4EDDA}84.93$_{\pm 1.29}$ & \cellcolor[HTML]{D4EDDA}{\ul 85.32$_{\pm 1.94}$} & \cellcolor[HTML]{D4EDDA}88.46$_{\pm 0.65}$ \\
 & \cellcolor[HTML]{EFEFEF}L-MAD-voting & \cellcolor[HTML]{D4EDDA}91.36$_{\pm 1.73}$ & \cellcolor[HTML]{D4EDDA}80.56$_{\pm 2.91}$ & \cellcolor[HTML]{D4EDDA}82.41$_{\pm 0.97}$ & \cellcolor[HTML]{D4EDDA}84.78$_{\pm 3.24}$ \\ \midrule
Qwen3-32B & Zeroshot & $91.87_{\pm 0.70}$ & $88.58_{\pm 1.58}$ & \textbf{87.16}$_{\pm 0.00}$ & $89.20_{\pm 0.76}$ \\
 & IRAC & \underline{94.31}$_{\pm 1.86}$ & $87.21_{\pm 0.79}$ & $85.32_{\pm 1.83}$ & $88.71_{\pm 1.50}$ \\
 & Fewshot & $92.68_{\pm 0.85}$ & $84.93_{\pm 1.32}$ & $81.65_{\pm 0.98}$ & $86.42_{\pm 1.07}$ \\
 & Self-Consistency & \textbf{95.12} & 87.67 & \textbf{87.16} & \textbf{89.98} \\
 & \cellcolor[HTML]{EFEFEF}L-MAD-consensus & \cellcolor[HTML]{D4EDDA}{\ul 92.68$_{\pm 3.25}$} & \cellcolor[HTML]{F8D7DA}87.67$_{\pm 4.31}$ & \cellcolor[HTML]{F8D7DA}{\ul 85.32$_{\pm 0.86}$} & \cellcolor[HTML]{F8D7DA}88.56$_{\pm 2.59}$ \\
 & \cellcolor[HTML]{EFEFEF}L-MAD-voting & \cellcolor[HTML]{D4EDDA}{\ul 92.68$_{\pm 0.00}$} & \cellcolor[HTML]{F8D7DA}82.19$_{\pm 1.73}$ & \cellcolor[HTML]{F8D7DA}80.73$_{\pm 4.84}$ & \cellcolor[HTML]{F8D7DA}85.20$_{\pm 1.94}$ \\ \midrule
Qwen3-8B & Zeroshot & $86.59_{\pm 0.00}$ & $81.74_{\pm 2.09}$ & $76.45_{\pm 1.40}$ & $81.59_{\pm 1.16}$ \\
 & IRAC & $88.62_{\pm 1.41}$ & $79.45_{\pm 0.00}$ & $79.20_{\pm 1.91}$ & $82.66_{\pm 1.11}$ \\
 & Fewshot & $90.24_{\pm 1.10}$ & $76.71_{\pm 1.45}$ & $74.31_{\pm 1.22}$ & $80.42_{\pm 1.27}$ \\
 & Self-Consistency & 87.80 & 79.45 & 81.65 & 82.97 \\
 & \cellcolor[HTML]{EFEFEF}L-MAD-consensus & \cellcolor[HTML]{D4EDDA}91.46$_{\pm 2.58}$ & \cellcolor[HTML]{D4EDDA}{\ul 90.41$_{\pm 1.94}$} & \cellcolor[HTML]{D4EDDA}79.82$_{\pm 0.65}$ & \cellcolor[HTML]{D4EDDA}87.23$_{\pm 1.29}$ \\
 & \cellcolor[HTML]{EFEFEF}L-MAD-voting & \cellcolor[HTML]{D4EDDA}91.46$_{\pm 0.86}$ & \cellcolor[HTML]{D4EDDA}\textbf{93.15$_{\pm 2.58}$ } & \cellcolor[HTML]{D4EDDA}83.49$_{\pm 1.10}$ & \cellcolor[HTML]{D4EDDA}{\ul 89.37$_{\pm 1.45}$} \\ \midrule
Llama3.1-8B & Zeroshot & $79.27_{\pm 0.00}$ & $71.23_{\pm 0.00}$ & $64.83_{\pm 0.53}$ & $71.88_{\pm 0.18}$ \\
 & IRAC & $79.67_{\pm 0.70}$ & $71.23_{\pm 0.00}$ & $68.81_{\pm 0.00}$ & $73.10_{\pm 0.23}$ \\
 & Fewshot & $63.41_{\pm 1.55}$ & $64.38_{\pm 1.12}$ & $62.39_{\pm 1.38}$ & $63.39_{\pm 1.36}$ \\
 & Self-Consistency & 79.27 & 71.23 & 68.81 & 73.10 \\
 & \cellcolor[HTML]{EFEFEF}L-MAD-consensus & \cellcolor[HTML]{F8D7DA}70.73$_{\pm 1.22}$ & \cellcolor[HTML]{F8D7DA}68.49$_{\pm 1.15}$ & \cellcolor[HTML]{F8D7DA}54.13$_{\pm 0.95}$ & \cellcolor[HTML]{F8D7DA}64.53$_{\pm 1.38}$ \\
 & \cellcolor[HTML]{EFEFEF}L-MAD-voting & \cellcolor[HTML]{F8D7DA}71.95$_{\pm 2.91}$ & \cellcolor[HTML]{F8D7DA}68.49$_{\pm 1.58}$ & \cellcolor[HTML]{D4EDDA}65.14$_{\pm 0.70}$ & \cellcolor[HTML]{F8D7DA}68.53$_{\pm 1.83}$ \\ \bottomrule
\end{tabular}
\end{table*}

\subsection{Performance of Multi-Agent Debate framework}

Table \ref{tab:main_results} presents the efficacy of the L-MAD framework, which scales non-linearly with the inherent competence of the base language model. Rather than providing a universal performance boost, L-MAD acts as a cognitive amplifier that reveals three distinct behavioral regimes.

First, for highly capable but non-saturating models (such as Qwen3-8B and Qwen3-30B), L-MAD serves as a powerful corrective mechanism. On average, the consensus and voting protocols outperform robust single-agent baselines like self-consistency and IRAC. In this "sweet spot," the debate structure successfully aggregates diverse reasoning paths, allowing agents to fact-check one another and converge on more nuanced legal entailments.

Second, we observe a ceiling effect with the most advanced architecture (Qwen3-32B). As a model's standalone reasoning approaches state-of-the-art, advanced single-agent prompting—particularly self-consistency—saturates the task performance, matching or even marginally exceeding the multi-agent setup. This suggests that when a model possesses sufficiently robust internal semantic representations of legal logic, the computational overhead and complexity of multi-agent collaboration may offer diminishing returns.

Finally, the results expose a critical ``competence threshold'' for multi-agent systems. When deployed on less-capable models (e.g., Llama3.1-8B), the L-MAD protocols actively degrade performance relative to isolated single-agent baselines. From a theoretical standpoint, this indicates that without a minimum baseline of deductive reasoning, multi-agent frameworks become vulnerable to collaborative hallucination. Instead of correcting mistakes, weak agents blindly validate each other's flawed premises, cascading errors, and forming a confidently incorrect echo chamber.





\subsection{Performance of Decision Protocols}

Comparing consensus and voting mechanisms reveals a clear trade-off between individual agent independence and collective alignment. Consensus protocols require agents to actively negotiate and converge on a single shared conclusion across debate rounds. In contrast, voting protocols preserve independent, parallel reasoning paths, delaying aggregation until the final step.

Our empirical results indicate that the optimal decision protocol depends heavily on the reasoning capacity of the underlying model. For larger models (e.g., 30B+ parameters), consensus protocols generally outperform voting. Because these models possess stronger internal logical consistency, they can engage in constructive evaluation—effectively identifying errors in competing arguments, correcting them, and synthesizing a more accurate final rationale. In this regime, the forced alignment mechanism functions as a rigorous, iterative peer-review process.

Conversely, for smaller models (e.g., 8B parameters), the voting mechanism significantly outperforms the consensus approach. Forced consensus in mid-tier models frequently induces sycophancy or premature convergence, where agents tend to uncritically adopt the initial or dominant reasoning path, even if it is fundamentally flawed. Voting mitigates this vulnerability by maintaining the independence of each agent's analysis. By delaying aggregation to the final stage, voting successfully leverages a "wisdom of the crowd" effect, preventing the final verdict from being derailed by a single, incorrect agent during the discussion phase.

\subsection{Number of Agents and Discussion Rounds}
\begin{figure*}[t]
    \centering
    \includegraphics[width=1\linewidth]{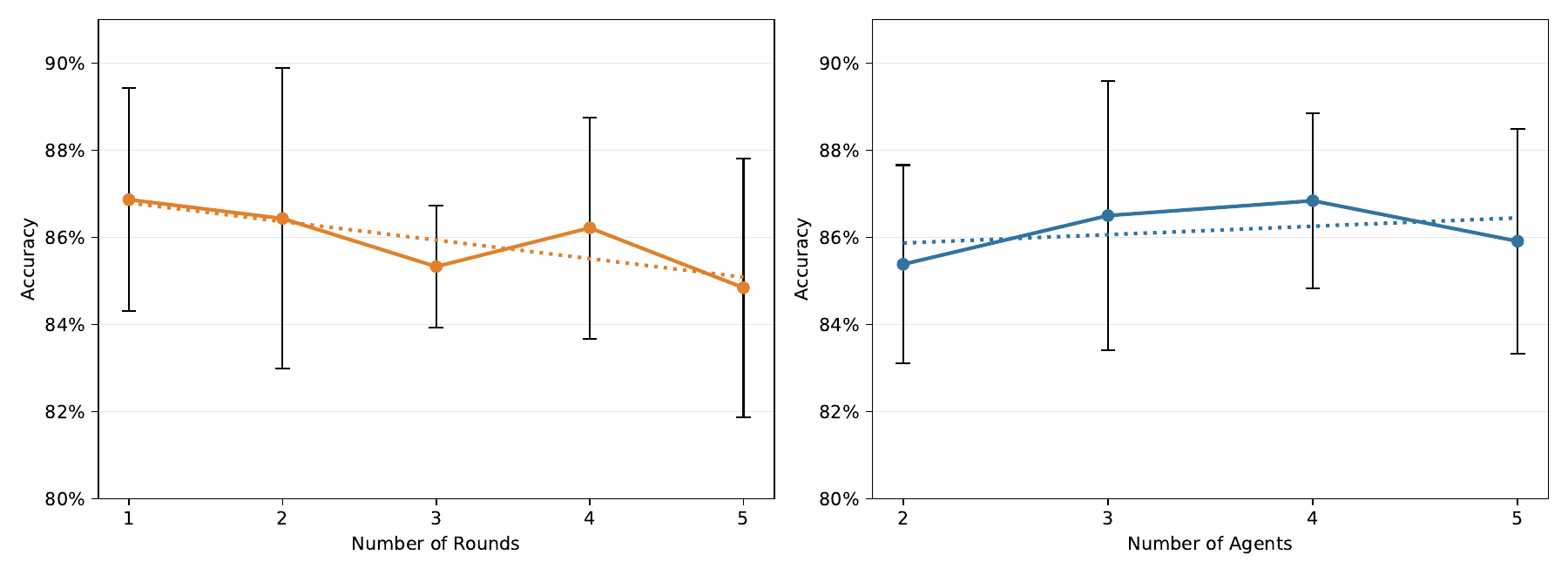}
    \caption{\textbf{Left}: System accuracy as a function of the number of deliberation rounds prior to voting. \textbf{Right}: System accuracy as a function of the number of participating agents. The final answers are derived using the ranked voting decision protocol. Trends are indicated by linear regression fits (dotted lines).}
    \label{fig:rounds_agents}
\end{figure*}

Prior literature presents conflicting hypotheses on the scaling properties of multi-agent debate. While some studies suggest that scaling both the number of participating agents and deliberation rounds monotonically improves performance via test-time computational scaling \cite{yin-etal-2023-exchange, wang-etal-2024-rethinking-bounds}, others warn that extended discourse may induce topic drift or echo-chamber effects \cite{becker2024multi}. To optimize the compute-accuracy trade-off and rigorously evaluate these claims within the domain of legal reasoning, we systematically ablate the multi-agent topology parameter space.

Specifically, we independently scale the agent population size ($N \in \{2, 3, 4, 5\}$) and the deliberation depth constraint ($T \in \{1, 2, 3, 4, 5\}$) on the Qwen3-8B backbone using the ranked-voting protocol. To isolate the marginal effect of each dimension, accuracy with respect to the number of rounds is averaged across all agent configurations, and conversely, accuracy with respect to the agent count is marginalized across all round limits.

The results reveal a clear divergence in how these two scaling axes govern task performance (Figure \ref{fig:rounds_agents}). Broadening the agent population yields a modest upward trend in accuracy (Figure \ref{fig:rounds_agents}, right), analogous to the variance reduction and coverage maximization observed in self-consistency techniques \cite{wang2022self}. Expanding the diversity of initial propositions generally benefits the final aggregation step by providing a broader sample of reasoning paths.

Conversely, scaling the deliberation depth exhibits a negative trajectory (Figure \ref{fig:rounds_agents}, left). Rather than self-refining toward the ground truth, agents subjected to extended discussion rounds demonstrate gradual performance degradation. This empirical finding challenges the assumed universal benefits of iterative self-correction \cite{NEURIPS2023_91edff07}, aligning instead with emerging skepticism regarding LLM self-refinement stability without external feedback \cite{huang2023large}. In specialized domains like legal entailment, extended deliberation intuitively amplifies the risk of adversarial drift, where agents over-critique valid logic and jointly regress toward suboptimal hypotheses.

\subsection{Qualitative Analysis}

To better understand the quantitative results, we conducted a systematic review of the debate transcripts across both successful and failed entailment cases. This evaluation categorizes the distinct behaviors of multi-agent reasoning into its fundamental strengths and practical limitations within rule-based environments.

\subsubsection{Advantages of Multi-Agent Debate}

\paragraph{Distributed Interpretation and Error Correction.}
In highly capable models (e.g., Qwen3-8B and Qwen3-30B), performance improvements are driven by substantive corrections rather than simple output aggregation. When multi-agent systems correct a unanimous failure of a single agent, the success typically stems from dividing the analytical effort. Figure~\ref{fig:correction_example} illustrates this on a complex statute. A single agent often focuses prematurely on a primary clause and misses critical exceptions. In contrast, assigning different roles allows agents to independently parse various parts of the legal text, enabling one agent to identify an overlooked exception and correct the group's understanding.

\begin{figure}[t]
\centering
\begin{tcolorbox}[colback=white, colframe=black!40, boxrule=0.4pt, arc=1pt, left=6pt, right=6pt, top=5pt, bottom=5pt]
\footnotesize
\textbf{Article 102} \quad Gold: Yes \quad Baseline: No/No/No \quad MAD: Yes\\[3pt]
\textbf{Statement.} An act performed by a restricted-capacity person as legal agent for another restricted-capacity person may be cancelled on grounds of incapacity.\\[2pt]
\textbf{Article.} An act by a restricted-capacity person as agent cannot be cancelled on grounds of capacity; provided, however, that this does not apply when acting as legal agent for another restricted-capacity person.\\[5pt]
Baseline (R1, R3): Reads only the main clause. The act cannot be cancelled. No\\[3pt]
Agent 1 (R1): The exception clause carves out an exception that applies here. Yes\\
Agent 2 (R1): Agree. The exception explicitly allows cancellation in this case. Yes\\
Agent 3 (R1): Agree. Input matches the exception. Yes\\[3pt]
\textbf{Final:} Yes (correct). A clause that zeroshot reads past is surfaced by the multi-agent parse.
\end{tcolorbox}
\caption{MAD-win on Civil Code Article 102. The zeroshot baseline misses the statutory exception in all three runs; L-MAD identifies it on the first turn.}
\label{fig:correction_example}
\end{figure}

\paragraph{Multi-Step Reasoning Convergence.}
Building on the ability to distribute reading, the debate framework also effectively connects multiple legal provisions. The multi-agent setup excels when a legal outcome requires combining two distinct articles (e.g., Articles 253 and 254 in Figure~\ref{fig:multistep_example}). While a single agent might stop after reading only one provision, collaborating agents can sequentially build upon each other's points to bridge logical gaps. This behavior constructs a chain of reasoning that single-pass prompting often fails to produce, which explains the heightened performance on datasets with complex legislative requirements.

\begin{figure}[t]
\centering
\begin{tcolorbox}[colback=white, colframe=black!40, boxrule=0.4pt, arc=1pt, left=6pt, right=6pt, top=5pt, bottom=5pt]
\footnotesize
\textbf{Articles 253 \& 254} \quad Gold: Yes \quad Baseline: No/No/No \quad MAD: Yes\\[3pt]
\textbf{Statement.} Co-owner A paid management costs. Co-owner B then transferred their share to C. A may claim reimbursement from C proportional to C's share.\\[2pt]
\textbf{Article 253.} Each co-owner pays management costs in proportion to their share.\\
\textbf{Article 254.} A claim that a co-owner holds against another co-owner may also be exercised against a successor in interest.\\[5pt]
Baseline (R1, R3): Considers only Article 253. Successor's liability is unclear. No\\[3pt]
Agent 1 (R1): Article 253 imposes proportional liability; Article 254 transmits the claim to successors. C inherits the obligation. Yes\\
Agent 2 (R1): Agree. The two articles together support the input. Yes\\
Agent 3 (R1): Agree. The chain is consistent with the statutory text. Yes\\[3pt]
\textbf{Final:} Yes (correct). The two-article chain is resolved on the first turn of debate.
\end{tcolorbox}
\caption{MAD-win on co-ownership reimbursement (Articles 253 and 254). Zeroshot fixates on a single article; L-MAD integrates both provisions in one turn.}
\label{fig:multistep_example}
\end{figure}

\paragraph{Voting Divergence as a Reliability Signal.}
Evaluating the voting framework reveals that disagreement among agents serves as a practical, natural indicator of task difficulty. For mid-tier models, instances with unanimous votes achieved approximately 70\% accuracy, while split votes—making up about 25\% of queries—dropped to 48\% accuracy. Protocols that force consensus obscure this metric by requiring all agents to align. By allowing agents to reason independently until the final decision, voting frameworks naturally expose when a model has reached the limits of its understanding. In real-world applications, this allows administrators to automatically route uncertain, split decisions to human experts while confidently automating unanimous cases.

\subsubsection{Failure Cases of Multi-Agent Debate}

\paragraph{Collaborative Hallucination and the Over-Qualification Cascade.}
For less capable models, such as Llama3.1-8B, the debate format introduces a compounding failure mode where errors are validated rather than corrected. As shown in Figure~\ref{fig:cascade_example}, this cascade begins when an agent introduces a flawed interpretation or an incorrect hesitation. Because subsequent agents lack the reasoning capacity to recognize the error, they adopt the incorrect premise and treat it as a factual constraint. Our analysis shows that expressions of uncertainty or hesitation appear ten times more frequently in the failure logs of weaker models compared to capable ones. This indicates that without a strong foundational reasoning ability, multi-agent frameworks can act as echo chambers that reinforce mistakes.

\begin{figure}[t]
\centering
\begin{tcolorbox}[colback=gray!5, colframe=gray!40, boxrule=0.5pt, arc=2pt, left=6pt, right=6pt, top=6pt, bottom=6pt]
\small
\textbf{Statement:} The benefit of a deadline cannot be waived when waiver harms the other party.\\
\textbf{Article 136(2):} The benefit may be waived; provided, however, that the waiver may not harm the other party. \quad \textbf{Gold:} No.\\[3pt]
\textbf{Agent 1 (round 1):} The statement reflects the exception in Article 136(2), which prohibits waiver when it harms the other party. Yes.\\[2pt]
\textbf{Agent 2 (round 1):} \textsc{Agree}. The input mirrors the exception clause. Yes.\\[2pt]
\textbf{Agent 3 (round 1):} \textsc{Agree}. Direct restatement of the exception. Yes.\\[2pt]
All round-2 agents reaffirm without disagreement.\\[3pt]
\textbf{Final:} Yes (incorrect). The article restricts the effect of waiver, not the act itself.
\end{tcolorbox}
\caption{Cascade on Civil Code Article 136(2). A literal misreading of the exception clause is adopted unchanged through the three remaining turns of the five-turn debate.}
\label{fig:cascade_example}
\end{figure}

\paragraph{Over-Deliberation Drift from Extended Debate}
A key limitation of allocating more discussion rounds is the risk of continuous performance degradation, previously identified as over-deliberation drift. Because the framework instructs agents to critique previous answers, extending the discussion selectively encourages the creation of unnecessary doubts. Figure~\ref{fig:flip_example} demonstrates how a correct initial voting is overturned in the second round by a flawed objection, leading the group away from the correct answer. When agents cannot resolve ambiguity, they tend to generate increasingly defensive responses instead of producing new insights. Therefore, implementing a rule to stop the debate early when unanimous agreement is reached is essential to preserve correct conclusions and prevent topic drift.

\begin{figure}[t]
\centering
\begin{tcolorbox}[colback=gray!5, colframe=gray!40, boxrule=0.5pt, arc=2pt, left=6pt, right=6pt, top=6pt, bottom=6pt]
\small
\textbf{Statement:} The pledgee of immovable property bears management costs unless the contract provides otherwise.\\
\textbf{Article 357:} The pledgee shall pay management costs and other burdens related to the immovable. \quad \textbf{Gold:} Yes.\\[3pt]
\textbf{Round 1 (all three agents):} The statement aligns with Article 357's assignment of costs to the pledgee. Yes.\\[2pt]
\textbf{Agent 1 (round 2):} \textsc{Disagree}. The clause unless provided otherwise is not in Article 357. No.\\[2pt]
\textbf{Agent 2 (round 2):} \textsc{Disagree}. The input introduces an exception not present in the article. No.\\[2pt]
\textbf{Agent 3 (round 2):} \textsc{Disagree}. Same reasoning. No.\\[3pt]
\textbf{Final:} No (incorrect). A correct round-1 voting is inverted by a second-round critique.
\end{tcolorbox}
\caption{Over-deliberation drift on Civil Code Article 357. Round 1 produces the correct answer; round 2 introduces a spurious objection that flips the entire group.}
\label{fig:flip_example}
\end{figure}

\paragraph{Fundamental Knowledge Boundaries.}
Despite these collaborative benefits, the framework consistently struggles with an irreducible subset of queries (roughly 17-24\%). Manual inspection shows that most of these universally failed cases require legal knowledge that is not directly stated in the supplied text. Because all participating agents share the same underlying model and knowledge constraints, they are equally unable to bridge these informational gaps. Consequently, the agents reach the same incorrect conclusion without productive debate. This highlights that simply increasing the amount of computation during inference is insufficient; multi-agent systems must be supported by external knowledge sources—such as legal precedents and hierarchical document context—to surpass this inherent performance limit.

\section{Related Work}

\paragraph{Legal NLP.}
Legal NLP benchmarks such as LexGLUE \citep{chalkidis-etal-2022-lexglue}, LegalBench \citep{NEURIPS2023_89e44582}, and LEXam \citep{fan2026lexam} show that legal language understanding requires models to handle domain-specific terminology, long contexts, and rule-based inference. Our work focuses on COLIEE Task~4~\citep{rabelo2024coliee, goebel2026coliee}, a binary legal textual entailment task over Japanese Civil Code articles. Prior COLIEE systems have explored retrieval, transformer-based entailment models, data augmentation, hybrid symbolic-neural methods, and Japanese LLM prompting for Task~4 \citep{10.1007/978-981-97-3076-6_8, steging2024hybrid, onaga2026kis}. More broadly, prompting methods such as in-context learning \citep{brown2020language}, chain-of-thought prompting \citep{wei2022chain}, self-consistency \citep{wang2022self}, and IRAC-style legal prompting \citep{yu2022legal} have been used to improve reasoning without model fine-tuning. However, iterative reasoning is not always reliable: self-refinement can help in some settings \citep{madaan2023self}, but LLMs often fail to self-correct without external feedback \citep{huang2023large}. In contrast to prior single-agent legal entailment systems, we study whether the final legal decision can be improved through multi-agent deliberation.

\paragraph{Multi-agent Frameworks.}
Multi-agent debate uses multiple LLM instances to propose, critique, and revise answers before aggregation. Prior work shows that such interaction can improve factuality and reasoning \citep{10.5555/3692070.3692537}, and communication-based frameworks study topologies such as memory, relay, report, and debate \citep{yin-etal-2023-exchange}. However, these gains are not universal: strong single-agent prompting can often match multi-agent discussion \citep{wang-etal-2024-rethinking-bounds}, and voting may explain a large part of the improvement attributed to debate \citep{choi2026debate}. MALLM frames multi-agent systems as a design space involving personas, response generators, communication paradigms, and decision protocols \citep{becker-etal-2025-mallm}. Persona prompting can diversify reasoning paths but may also introduce instability \citep{kim2024persona}. Aggregation protocols also matter: consensus forces agents to converge, while voting preserves independent judgments until the final decision \citep{yang2024llm, kaesberg-etal-2025-voting}. Our work connects these ideas to statute-grounded legal entailment, comparing consensus and voting across several LLM backbones and analyzing when debate corrects errors, over-deliberates, or amplifies flawed reasoning.


\section{Conclusion}

In this work, we introduced L-MAD, a framework to evaluate multi-agent debate for complex legal reasoning. We found that simply adding more agents or discussion rounds does not guarantee better results. Instead, MAD acts as a performance multiplier that is strictly limited by the base model's capabilities: while it helps capable models collaborate and correct mistakes, it offers little benefit to state-of-the-art models and causes weaker models to reinforce each other's errors.
Furthermore, we demonstrated that forcing models to debate for too long actually harms accuracy, as agents begin to second-guess correct answers. Our results suggest that to improve performance, it is better to increase the number of diverse agents voting independently than to force extended, consensus-driven discussions.

These findings are crucial for safely deploying language models in high-stakes environments like law. When agents naturally disagree during a vote, it serves as a reliable signal to route the difficult problem to a human expert. Ultimately, effective multi-agent collaboration requires carefully balancing the capabilities of the models and the structure of the debate, rather than assuming that longer discussions will naturally yield better answers.

For future work, we plan to explore how asymmetric power dynamics and hierarchical relationships among agents influence debate outcomes. Specifically, investigating whether authoritative or collaborative leadership personas are more effective at guiding discussions could yield more robust decision protocols. Additionally, integrating Explainable AI (XAI) techniques into the MAD framework should be explored to further enhance the transparency and reliability of LLMs in high-stakes legal applications.





\section*{Acknowledgements}
We sincerely thank the anonymous reviewers for their insightful comments and valuable feedback, which helped improve the quality of this manuscript. This work was partly supported by the Japan Science and Technology Agency (JST) as part of the Adopting Sustainable Partnerships for Innovative Research Ecosystem (ASPIRE), Grant Number JPMJAP25B2.



\section*{Impact Statement}

This paper presents work whose primary goal is to advance the reliability and interpretability of Machine Learning systems in high-stakes, knowledge-dense domains, specifically focusing on legal reasoning. As Large Language Models are increasingly integrated into judicial, legal, and regulatory workflows, ensuring their deductive accuracy and epistemological robustness is of paramount societal importance. 

Our research highlights both the potential and the systemic risks of deploying multi-agent systems in these environments. On the positive side, our findings regarding vote disagreement provide a native, training-free confidence signal. This enables the design of safe, hybrid human-in-the-loop systems where high-uncertainty cases are reliably routed to human legal professionals, thereby mitigating the risk of fully autonomous legal malpractice.

\nocite{langley00}

\bibliography{example_paper}

@inproceedings{langley00,
 author    = {P. Langley},
 title     = {Crafting Papers on Machine Learning},
 year      = {2000},
 pages     = {1207--1216},
 editor    = {Pat Langley},
 booktitle     = {Proceedings of the 17th International Conference
              on Machine Learning (ICML 2000)},
 address   = {Stanford, CA},
 publisher = {Morgan Kaufmann}
}

@article{yu2022legal,
  title={Legal prompting: Teaching a language model to think like a lawyer},
  author={Yu, Fangyi and Quartey, Lee and Schilder, Frank},
  journal={arXiv preprint arXiv:2212.01326},
  year={2022}
}

@article{wang2022self,
  title={Self-consistency improves chain of thought reasoning in language models},
  author={Wang, Xuezhi and Wei, Jason and Schuurmans, Dale and Le, Quoc and Chi, Ed and Narang, Sharan and Chowdhery, Aakanksha and Zhou, Denny},
  journal={arXiv preprint arXiv:2203.11171},
  year={2022}
}

@article{qwen3,
    title={Qwen3 Technical Report}, 
    author={An Yang and Anfeng Li and Baosong Yang and Beichen Zhang and Binyuan Hui and Bo Zheng and Bowen Yu and Chang Gao and Chengen Huang and Chenxu Lv and Chujie Zheng and Dayiheng Liu and Fan Zhou and Fei Huang and Feng Hu and Hao Ge and Haoran Wei and Huan Lin and Jialong Tang and Jian Yang and Jianhong Tu and Jianwei Zhang and Jianxin Yang and Jiaxi Yang and Jing Zhou and Jingren Zhou and Junyang Lin and Kai Dang and Keqin Bao and Kexin Yang and Le Yu and Lianghao Deng and Mei Li and Mingfeng Xue and Mingze Li and Pei Zhang and Peng Wang and Qin Zhu and Rui Men and Ruize Gao and Shixuan Liu and Shuang Luo and Tianhao Li and Tianyi Tang and Wenbiao Yin and Xingzhang Ren and Xinyu Wang and Xinyu Zhang and Xuancheng Ren and Yang Fan and Yang Su and Yichang Zhang and Yinger Zhang and Yu Wan and Yuqiong Liu and Zekun Wang and Zeyu Cui and Zhenru Zhang and Zhipeng Zhou and Zihan Qiu},
    journal = {arXiv preprint arXiv:2505.09388},
    year={2025}
}

@article{dubey2024llama,
  title={The llama 3 herd of models},
  author={Dubey, Abhimanyu and Jauhri, Abhinav and Pandey, Abhinav and Kadian, Abhishek and Al-Dahle, Ahmad and Letman, Aiesha and Mathur, Akhil and Schelten, Alan and Yang, Amy and Fan, Angela and others},
  journal={arXiv e-prints},
  pages={arXiv--2407},
  year={2024}
}

@article{huang2023large,
 title={Large language models cannot self-correct reasoning yet},
 author={Huang, Jie and Chen, Xinyun and Mishra, Swaroop and Zheng, Huaixiu Steven and Yu, Adams Wei and Song, Xinying and Zhou, Denny},
 journal={arXiv preprint arXiv:2310.01798},
 year={2023}
}

@inproceedings{yin-etal-2023-exchange,
    title = "Exchange-of-Thought: Enhancing Large Language Model Capabilities through Cross-Model Communication",
    author = "Yin, Zhangyue  and
      Sun, Qiushi  and
      Chang, Cheng  and
      Guo, Qipeng  and
      Dai, Junqi  and
      Huang, Xuanjing  and
      Qiu, Xipeng",
    editor = "Bouamor, Houda  and
      Pino, Juan  and
      Bali, Kalika",
    booktitle = "Proceedings of the 2023 Conference on Empirical Methods in Natural Language Processing",
    month = dec,
    year = "2023",
    address = "Singapore",
    publisher = "Association for Computational Linguistics",
    url = "https://aclanthology.org/2023.emnlp-main.936/",
    doi = "10.18653/v1/2023.emnlp-main.936",
    pages = "15135--15153"
}

@inproceedings{10.5555/3692070.3692537,
author = {Du, Yilun and Li, Shuang and Torralba, Antonio and Tenenbaum, Joshua B. and Mordatch, Igor},
title = {Improving factuality and reasoning in language models through multiagent debate},
year = {2024},
publisher = {JMLR.org},
booktitle = {Proceedings of the 41st International Conference on Machine Learning},
articleno = {467},
numpages = {31},
location = {Vienna, Austria},
series = {ICML'24}
}

@article{becker2024multi,
  title={Multi-agent large language models for conversational task-solving},
  author={Becker, Jonas},
  journal={arXiv preprint arXiv:2410.22932},
  year={2024}
}

@inproceedings{wang-etal-2024-rethinking-bounds,
    title = "Rethinking the Bounds of {LLM} Reasoning: Are Multi-Agent Discussions the Key?",
    author = "Wang, Qineng  and
      Wang, Zihao  and
      Su, Ying  and
      Tong, Hanghang  and
      Song, Yangqiu",
    editor = "Ku, Lun-Wei  and
      Martins, Andre  and
      Srikumar, Vivek",
    booktitle = "Proceedings of the 62nd Annual Meeting of the Association for Computational Linguistics (Volume 1: Long Papers)",
    month = aug,
    year = "2024",
    address = "Bangkok, Thailand",
    publisher = "Association for Computational Linguistics",
    url = "https://aclanthology.org/2024.acl-long.331/",
    doi = "10.18653/v1/2024.acl-long.331",
    pages = "6106--6131"
}

@inproceedings{NEURIPS2023_91edff07,
 author = {Madaan, Aman and Tandon, Niket and Gupta, Prakhar and Hallinan, Skyler and Gao, Luyu and Wiegreffe, Sarah and Alon, Uri and Dziri, Nouha and Prabhumoye, Shrimai and Yang, Yiming and Gupta, Shashank and Majumder, Bodhisattwa Prasad and Hermann, Katherine and Welleck, Sean and Yazdanbakhsh, Amir and Clark, Peter},
 booktitle = {Advances in Neural Information Processing Systems},
 editor = {A. Oh and T. Naumann and A. Globerson and K. Saenko and M. Hardt and S. Levine},
 pages = {46534--46594},
 publisher = {Curran Associates, Inc.},
 title = {Self-Refine: Iterative Refinement with Self-Feedback},
 volume = {36},
 year = {2023}
}

@inproceedings{yang2024llm,
  title={Llm voting: Human choices and ai collective decision-making},
  author={Yang, Joshua C and Dailisan, Damian and Korecki, Marcin and Hausladen, Carina I and Helbing, Dirk},
  booktitle={Proceedings of the AAAI/ACM Conference on AI, Ethics, and Society},
  volume={7},
  number={1},
  pages={1696--1708},
  year={2024}
}

@article{kim2024persona,
  title={Persona is a double-edged sword: Mitigating the negative impact of role-playing prompts in zero-shot reasoning tasks},
  author={Kim, Junseok and Yang, Nakyeong and Jung, Kyomin},
  journal={arXiv preprint arXiv:2408.08631},
  year={2024}
}

@article{brown2020language,
  title={Language models are few-shot learners},
  author={Brown, Tom and Mann, Benjamin and Ryder, Nick and Subbiah, Melanie and Kaplan, Jared D and Dhariwal, Prafulla and Neelakantan, Arvind and Shyam, Pranav and Sastry, Girish and Askell, Amanda and others},
  journal={Advances in neural information processing systems},
  volume={33},
  pages={1877--1901},
  year={2020}
}

@inproceedings{chalkidis-etal-2022-lexglue,
    title = "{L}ex{GLUE}: A Benchmark Dataset for Legal Language Understanding in {E}nglish",
    author = "Chalkidis, Ilias  and
      Jana, Abhik  and
      Hartung, Dirk  and
      Bommarito, Michael  and
      Androutsopoulos, Ion  and
      Katz, Daniel  and
      Aletras, Nikolaos",
    editor = "Muresan, Smaranda  and
      Nakov, Preslav  and
      Villavicencio, Aline",
    booktitle = "Proceedings of the 60th Annual Meeting of the Association for Computational Linguistics (Volume 1: Long Papers)",
    month = may,
    year = "2022",
    address = "Dublin, Ireland",
    publisher = "Association for Computational Linguistics",
    url = "https://aclanthology.org/2022.acl-long.297/",
    doi = "10.18653/v1/2022.acl-long.297",
    pages = "4310--4330"
}

@inproceedings{NEURIPS2023_89e44582,
 author = {Guha, Neel and Nyarko, Julian and Ho, Daniel and R\'{e}, Christopher and Chilton, Adam and K, Aditya and Chohlas-Wood, Alex and Peters, Austin and Waldon, Brandon and Rockmore, Daniel and Zambrano, Diego and Talisman, Dmitry and Hoque, Enam and Surani, Faiz and Fagan, Frank and Sarfaty, Galit and Dickinson, Gregory and Porat, Haggai and Hegland, Jason and Wu, Jessica and Nudell, Joe and Niklaus, Joel and Nay, John and Choi, Jonathan and Tobia, Kevin and Hagan, Margaret and Ma, Megan and Livermore, Michael and Rasumov-Rahe, Nikon and Holzenberger, Nils and Kolt, Noam and Henderson, Peter and Rehaag, Sean and Goel, Sharad and Gao, Shang and Williams, Spencer and Gandhi, Sunny and Zur, Tom and Iyer, Varun and Li, Zehua},
 booktitle = {Advances in Neural Information Processing Systems},
 editor = {A. Oh and T. Naumann and A. Globerson and K. Saenko and M. Hardt and S. Levine},
 pages = {44123--44279},
 publisher = {Curran Associates, Inc.},
 title = {LegalBench: A Collaboratively Built Benchmark for Measuring Legal Reasoning in Large Language Models},
 volume = {36},
 year = {2023}
}

@article{wei2022chain,
  title={Chain-of-thought prompting elicits reasoning in large language models},
  author={Wei, Jason and Wang, Xuezhi and Schuurmans, Dale and Bosma, Maarten and Xia, Fei and Chi, Ed and Le, Quoc V and Zhou, Denny and others},
  journal={Advances in neural information processing systems},
  volume={35},
  pages={24824--24837},
  year={2022}
}

@article{madaan2023self,
  title={Self-refine: Iterative refinement with self-feedback},
  author={Madaan, Aman and Tandon, Niket and Gupta, Prakhar and Hallinan, Skyler and Gao, Luyu and Wiegreffe, Sarah and Alon, Uri and Dziri, Nouha and Prabhumoye, Shrimai and Yang, Yiming and others},
  journal={Advances in neural information processing systems},
  volume={36},
  pages={46534--46594},
  year={2023}
}

@inproceedings{becker-etal-2025-mallm,
    title = "{MALLM}: Multi-Agent Large Language Models Framework",
    author = "Becker, Jonas  and
      Kaesberg, Lars Benedikt  and
      Bauer, Niklas  and
      Wahle, Jan Philip  and
      Ruas, Terry  and
      Gipp, Bela",
    editor = {Habernal, Ivan  and
      Schulam, Peter  and
      Tiedemann, J{\"o}rg},
    booktitle = "Proceedings of the 2025 Conference on Empirical Methods in Natural Language Processing: System Demonstrations",
    month = nov,
    year = "2025",
    address = "Suzhou, China",
    publisher = "Association for Computational Linguistics",
    url = "https://aclanthology.org/2025.emnlp-demos.29/",
    doi = "10.18653/v1/2025.emnlp-demos.29",
    pages = "418--439",
    ISBN = "979-8-89176-334-0"
}

@inproceedings{10.1007/978-981-97-3076-6_8,
author = {Goebel, Randy and Kano, Yoshinobu and Kim, Mi-Young and Rabelo, Juliano and Satoh, Ken and Yoshioka, Masaharu},
title = {Overview of Benchmark Datasets and Methods for the Legal Information Extraction/Entailment Competition (COLIEE) 2024},
year = {2024},
isbn = {978-981-97-3075-9},
publisher = {Springer-Verlag},
address = {Berlin, Heidelberg},
url = {https://doi.org/10.1007/978-981-97-3076-6_8},
doi = {10.1007/978-981-97-3076-6_8},
booktitle = {New Frontiers in Artificial Intelligence: JSAI International Symposium on Artificial Intelligence, JSAI-IsAI 2024, Hamamatsu, Japan, May 28–29, 2024, Proceedings},
pages = {109–124},
numpages = {16},
location = {Hamamatsu, Japan}
}

@article{aoki2022data,
  title={Data-augmentation method for bert-based legal textual entailment systems in coliee statute law task},
  author={Aoki, Yasuhiro and Yoshioka, Masaharu and Suzuki, Youta},
  journal={The Review of Socionetwork Strategies},
  volume={16},
  number={1},
  pages={175--196},
  year={2022},
  publisher={Springer}
}

@inproceedings{steging2024hybrid,
  title={A hybrid approach to legal textual entailment},
  author={Steging, Cor and Leeuwen, LV},
  booktitle={Proceedings of the Eighteenth International Workshop on Juris-Informatics (JURISIN 2024)},
  year={2024}
}

@article{rabelo2024coliee,
  author    = {Juliano Rabelo and Randy Goebel and Mi-Young Kim and Yoshinobu Kano and Masaharu Yoshioka and Ken Satoh},
  title     = {Overview and Discussion of the Competition on Legal Information Extraction/Entailment ({COLIEE}) 2023},
  journal   = {The Review of Socionetwork Strategies},
  volume    = {18},
  number    = {1},
  pages     = {27--47},
  year      = {2024}
}

@inproceedings{fan2026lexam,
  title={LEXam: Benchmarking legal reasoning on 340 law exams},
  author={Fan, Yu and Ni, Jingwei and Merane, Jakob and Salimbeni, Etienne and Tian, Yang and Hermstr{\"u}wer, Yoan and Huang, Yinya and Akhtar, Mubashara and Geering, Florian and Dreyer, Oliver and others},
  booktitle={International Conference on Learning Representations (ICLR)},
  year={2026}
}

@inproceedings{kaesberg-etal-2025-voting,
    title = "Voting or Consensus? Decision-Making in Multi-Agent Debate",
    author = "Kaesberg, Lars Benedikt  and
      Becker, Jonas  and
      Wahle, Jan Philip  and
      Ruas, Terry  and
      Gipp, Bela",
    editor = "Che, Wanxiang  and
      Nabende, Joyce  and
      Shutova, Ekaterina  and
      Pilehvar, Mohammad Taher",
    booktitle = "Findings of the Association for Computational Linguistics: ACL 2025",
    month = jul,
    year = "2025",
    address = "Vienna, Austria",
    publisher = "Association for Computational Linguistics",
    url = "https://aclanthology.org/2025.findings-acl.606/",
    doi = "10.18653/v1/2025.findings-acl.606",
    pages = "11640--11671",
    ISBN = "979-8-89176-256-5"
}

@article{choi2026debate,
  title={Debate or vote: Which yields better decisions in multi-agent large language models?},
  author={Choi, Hyeong Kyu and Zhu, Jerry and Li, Sharon},
  journal={Advances in Neural Information Processing Systems},
  volume={38},
  pages={101732--101764},
  year={2026}
}

@article{goebel2026coliee,
  title={The COLIEE 2025 Competition on Legal Information Extraction and Entailment: Overview, Discussion, and Dataset Expansion},
  author={Goebel, Randy and Kano, Yoshinobu and Kim, Mi-Young and Kwan, Calum and Rabelo, Juliano and Satoh, Ken and Yamada, Hiroaki and Yoshioka, Masaharu},
  journal={The Review of Socionetwork Strategies},
  pages={1--31},
  year={2026},
  publisher={Springer}
}

@article{onaga2026kis,
  title={Kis: Coliee 2025 task 4 solver using Japanese llm},
  author={Onaga, Takaaki and Kano, Yoshinobu},
  journal={The Review of Socionetwork Strategies},
  pages={1--19},
  year={2026},
  publisher={Springer}
}

@article{fu2022complexity,
  title={Complexity-based prompting for multi-step reasoning},
  author={Fu, Yao and Peng, Hao and Sabharwal, Ashish and Clark, Peter and Khot, Tushar},
  journal={arXiv preprint arXiv:2210.00720},
  year={2022}
}

@article{aletras2016predicting,
  title={Predicting judicial decisions of the European Court of Human Rights: A natural language processing perspective},
  author={Aletras, Nikolaos and Tsarapatsanis, Dimitrios and Preo{\c{t}}iuc-Pietro, Daniel and Lampos, Vasileios},
  journal={PeerJ computer science},
  volume={2},
  pages={e93},
  year={2016},
  publisher={PeerJ Inc.}
}

@inproceedings{zhong-etal-2020-nlp,
    title = "How Does {NLP} Benefit Legal System: A Summary of Legal Artificial Intelligence",
    author = "Zhong, Haoxi  and
      Xiao, Chaojun  and
      Tu, Cunchao  and
      Zhang, Tianyang  and
      Liu, Zhiyuan  and
      Sun, Maosong",
    editor = "Jurafsky, Dan  and
      Chai, Joyce  and
      Schluter, Natalie  and
      Tetreault, Joel",
    booktitle = "Proceedings of the 58th Annual Meeting of the Association for Computational Linguistics",
    month = jul,
    year = "2020",
    address = "Online",
    publisher = "Association for Computational Linguistics",
    url = "https://aclanthology.org/2020.acl-main.466/",
    doi = "10.18653/v1/2020.acl-main.466",
    pages = "5218--5230"
}

@article{bilgin2024exploring,
  title={Exploring prompting approaches in legal textual entailment},
  author={Bilgin, Onur and Fields, Logan and Laverghetta Jr, Antonio and Marji, Zaid and Nighojkar, Animesh and Steinle, Stephen and Licato, John},
  journal={The Review of Socionetwork Strategies},
  volume={18},
  number={1},
  pages={75--100},
  year={2024},
  publisher={Springer}
}

@inproceedings{chalkidis-etal-2020-legal,
    title = "{LEGAL}-{BERT}: The Muppets straight out of Law School",
    author = "Chalkidis, Ilias  and
      Fergadiotis, Manos  and
      Malakasiotis, Prodromos  and
      Aletras, Nikolaos  and
      Androutsopoulos, Ion",
    editor = "Cohn, Trevor  and
      He, Yulan  and
      Liu, Yang",
    booktitle = "Findings of the Association for Computational Linguistics: EMNLP 2020",
    month = nov,
    year = "2020",
    address = "Online",
    publisher = "Association for Computational Linguistics",
    url = "https://aclanthology.org/2020.findings-emnlp.261/",
    doi = "10.18653/v1/2020.findings-emnlp.261",
    pages = "2898--2904"
}

@article{xiao2021lawformer,
  title={Lawformer: A pre-trained language model for chinese legal long documents},
  author={Xiao, Chaojun and Hu, Xueyu and Liu, Zhiyuan and Tu, Cunchao and Sun, Maosong},
  journal={AI Open},
  volume={2},
  pages={79--84},
  year={2021},
  publisher={Elsevier}
}

@inproceedings{nie-etal-2020-adversarial,
    title = "Adversarial {NLI}: A New Benchmark for Natural Language Understanding",
    author = "Nie, Yixin  and
      Williams, Adina  and
      Dinan, Emily  and
      Bansal, Mohit  and
      Weston, Jason  and
      Kiela, Douwe",
    editor = "Jurafsky, Dan  and
      Chai, Joyce  and
      Schluter, Natalie  and
      Tetreault, Joel",
    booktitle = "Proceedings of the 58th Annual Meeting of the Association for Computational Linguistics",
    month = jul,
    year = "2020",
    address = "Online",
    publisher = "Association for Computational Linguistics",
    url = "https://aclanthology.org/2020.acl-main.441/",
    doi = "10.18653/v1/2020.acl-main.441",
    pages = "4885--4901"
}

@inproceedings{kaesberg2025voting,
  title={Voting or consensus? decision-making in multi-agent debate},
  author={Kaesberg, Lars Benedikt and Becker, Jonas and Wahle, Jan Philip and Ruas, Terry and Gipp, Bela},
  booktitle={Findings of the Association for Computational Linguistics: ACL 2025},
  pages={11640--11671},
  year={2025}
}
\bibliographystyle{icml2026}

\newpage
\appendix
\onecolumn
\section{Additional Experiment Details.}

\subsection{Agents' Personas}
\textbf{Judge}: You are an impartial, highly analytical Japanese Presiding Judge in a civil chamber. Your objective is to evaluate the arguments of the Plaintiff and Defendant, verify their statutory mapping against the Civil Code, and render a final verdict.

\textbf{Attorney}: You are a precise, methodical Japanese civil litigator. Your sole objective is to prove that the Premise legally ENTAILS the Hypothesis under the Japanese Civil Code.

\textbf{Lawyer}: You are a sharp, analytical Japanese civil lawyer. Your objective is to prove that the Premise does NOT legally entail the Hypothesis.

\subsection{Prompt Templates}
\label{app:prompts}

This section provides the exact prompt templates used for the baselines evaluated in our experiments. All experiments were conducted using Japanese prompts, natively aligning with the language of the Japanese Civil Code corpus. The text enclosed in brackets \texttt{\{...\}} represents dynamic variables substituted at inference time.

\subsubsection{Zeroshot Prompting}
\begin{CJK*}{UTF8}{ipxm}
\begin{tcolorbox}[colback=gray!5, colframe=gray!40, boxrule=0.5pt, arc=2pt, left=6pt, right=6pt, top=6pt, bottom=6pt, title=\textbf{System Prompt (Zeroshot)}]

\begin{verbatim}
あなたは日本の民法に精通した法律専門家です。
与えられた民法条文と法律問題を注意深く分析し、
条文が問題の答えを「はい」と裏付けているかどうかを判断してください。

回答は必ず以下の形式で出力してください：
判断: Y または N
理由: （簡潔な説明）
\end{verbatim}
\end{tcolorbox}

\begin{tcolorbox}[colback=white, colframe=black!40, boxrule=0.4pt, arc=1pt, left=6pt, right=6pt, top=5pt, bottom=5pt, title=\textbf{User Prompt (Zeroshot)}]
\begin{verbatim}
以下の民法条文と法律問題を読んでください。

【関連条文】
{article_text}

【法律問題】
{query}

この条文を根拠として、問題の答えは「はい」と言えますか？
判断: Y または N で答え、その理由を説明してください。
\end{verbatim}
\end{tcolorbox}

\subsubsection{IRAC (Legal Reasoning CoT) Prompting}
\begin{tcolorbox}[colback=gray!5, colframe=gray!40, boxrule=0.5pt, arc=2pt, left=6pt, right=6pt, top=6pt, bottom=6pt, title=\textbf{System Prompt (IRAC)}]
\begin{verbatim}
あなたは日本の民法に精通した法律専門家です。
提示された法的推論の手法に従い、仮説が「真」か「偽」かを分析してください。
手法：争点、法則、適用、結論。
回答は必ず以下の形式で出力してください：
判断: Y または N
理由: （簡潔な説明）
\end{verbatim}
\end{tcolorbox}

\begin{tcolorbox}[colback=white, colframe=black!40, boxrule=0.4pt, arc=1pt, left=6pt, right=6pt, top=5pt, bottom=5pt, title=\textbf{User Prompt (IRAC)}]
\begin{verbatim}
以下の民法条文と法律問題を読んでください。

【関連条文】
{article_text}

【法律問題】
{query}

この条文を根拠として、問題の答えは「はい」と言えますか？
判断: Y または N で答え、その理由を説明してください。
\end{verbatim}
\end{tcolorbox}

\subsubsection{Fewshot Prompting}
\begin{tcolorbox}[colback=gray!5, colframe=gray!40, boxrule=0.5pt, arc=2pt, left=6pt, right=6pt, top=6pt, bottom=6pt, title=\textbf{System Prompt (Fewshot)}]
\begin{verbatim}
あなたは日本の民法に精通した法律専門家です。
与えられた民法条文と法律問題を分析し、
条文が問題の答えを「はい」と裏付けているか判断してください。

回答は必ず以下の形式で出力してください：
判断: Y または N
理由: （簡潔な説明）
\end{verbatim}
\end{tcolorbox}

\begin{tcolorbox}[colback=white, colframe=black!40, boxrule=0.4pt, arc=1pt, left=6pt, right=6pt, top=5pt, bottom=5pt, title=\textbf{User Prompt (Fewshot)}]
\begin{verbatim}
以下の例を参考にして判断してください。

{examples}

今度はあなたが判断してください。

【関連条文】
{article_text}

【法律問題】
{query}

判断: Y または N で答え、その理由を説明してください。
\end{verbatim}
\end{tcolorbox}
\end{CJK*}



\end{document}